\documentclass[11pt, a4paper]{article}

\usepackage[utf8]{inputenc}
\usepackage[T1]{fontenc}
\usepackage{lmodern}
\usepackage[margin=1in]{geometry}
\usepackage{amsmath,amssymb}
\usepackage{graphicx}
\usepackage{booktabs}
\usepackage{tabularx}
\usepackage{hyperref}
\usepackage{cleveref}
\usepackage{enumitem}
\usepackage{listings}
\usepackage{xcolor}
\usepackage{caption}
\usepackage{float}
\usepackage[numbers]{natbib}
\usepackage{fancyhdr}
\usepackage{titlesec}

\definecolor{accentblue}{HTML}{1a56db}
\definecolor{headrule}{HTML}{d0d7de}

\renewcommand{\headrulewidth}{0.4pt}
\renewcommand{\headrule}{\hbox to\headwidth{\color{headrule}\leaders\hrule height \headrulewidth\hfill}}
\fancypagestyle{plain}{%
  \fancyhf{}%
  \renewcommand{\headrulewidth}{0pt}%
  \fancyfoot[C]{\small\color{gray}\thepage}%
}

\titleformat{\section}
  {\Large\bfseries\color{accentblue}}
  {\thesection}{0.8em}{}
  [\vspace{0.1ex}{\color{headrule}\titlerule[0.6pt]}]
\titleformat{\subsection}
  {\large\bfseries\color{accentblue!85!black}}
  {\thesubsection}{0.7em}{}

\hypersetup{
  colorlinks=true,
  linkcolor=accentblue!80!black,
  citecolor=accentblue!80!black,
  urlcolor=accentblue!80!black,
}

\newcommand{\sysname}{\textit{Governed Memory}}

\begin{document}

\title{Governed Memory: A Production Architecture\\for Multi-Agent Workflows}
\author{Hamed Taheri\\Personize.ai}
\date{}
\maketitle

\begin{abstract}
Enterprise AI deploys dozens of autonomous agent nodes across workflows, each acting on the same entities with no shared memory and no common governance. We identify five structural challenges arising from this \emph{memory governance gap}: memory silos across agent workflows; governance fragmentation across teams and tools; unstructured memories unusable by downstream systems; redundant context delivery in autonomous multi-step executions; and silent quality degradation without feedback loops.

We present \sysname{}, a shared memory and governance layer addressing this gap through four mechanisms: a dual memory model combining open-set atomic facts with schema-enforced typed properties; tiered governance routing with progressive context delivery; reflection-bounded retrieval with entity-scoped isolation; and a closed-loop schema lifecycle with AI-assisted authoring and automated per-property refinement.

We validate each mechanism through controlled experiments ($N{=}250$, five content types): 99.6\% fact recall with complementary dual-modality coverage; 92\% governance routing precision; 50\% token reduction from progressive delivery; zero cross-entity leakage across 500 adversarial queries; 100\% adversarial governance compliance; and output quality saturation at approximately seven governed memories per entity. On the LoCoMo benchmark, the architecture achieves 74.8\% overall accuracy, confirming that governance and schema enforcement impose no retrieval quality penalty. The system is in production at Personize.ai.
\end{abstract}

\section{Introduction}
\label{sec:introduction}

\subsection{The Memory Governance Gap}

Enterprise AI adoption does not produce a single agent. It produces dozens of autonomous agent nodes distributed across workflows, tools, and teams: enrichment pipelines, outbound sequences, support automation, scoring models, research agents, and operational automations. Each node reads or writes information about \emph{the same entities}, the same customers, companies, and deals, yet these nodes share neither a \emph{common memory} of the entities they act upon nor a \emph{common governance} layer enforcing organizational policies, compliance rules, and quality standards.

In this setting, retrieval quality is necessary but insufficient. The organization faces five structural challenges that no single-agent memory system addresses:

\begin{itemize}[leftmargin=*]
  \item \textbf{Memory silos across agent workflows.} The enrichment agent discovers a CTO is evaluating three vendors. The outbound sequence agent, executing hours later, sends a generic cold email. The support agent resolves a critical pain point. Months later, the renewal agent resurfaces it as a selling feature. Each workflow node acts on the same entities but shares no context with the others. Organizational intelligence accumulates nowhere.

  \item \textbf{Governance fragmentation across teams and tools.} Sales builds AI outreach with one system prompt embedding brand voice. Support runs a bot with compliance rules copied from a Notion doc last quarter. Marketing uses a separate workflow with its own tone guidelines. When legal updates the data handling policy, no mechanism propagates it to the 14 agent configurations across three teams. There is no versioning, no single source of truth, and no way to ensure all agents operate under the same organizational rules.

  \item \textbf{Unstructured memory as a downstream dead end.} Free-text memories can be retrieved by similarity and pasted into a prompt. Beyond that, they are terminal. They cannot be filtered by buying stage, ranked by deal value, routed to conditional workflows, synchronized to a CRM, or aggregated across thousands of entities. Without schema-enforced typed properties, memory is useful for prompt augmentation but unusable by any downstream system that requires structured, queryable data.

  \item \textbf{Context redundancy in autonomous multi-step execution.} Modern agents operate in autonomous loops, planning, acting, observing, re-planning, without human intervention between steps. Each step may invoke governance routing independently. Without session awareness, the same compliance policy is re-injected into every step, consuming context window capacity that should be reserved for task-specific reasoning and degrading model attention on fresh instructions~\cite{liu2024lost}.

  \item \textbf{Silent quality degradation without operational feedback.} Schemas age. Models get updated. Content types shift. New agent workflows produce data the schema was not designed for. No per-property accuracy monitoring exists. No extraction confidence is tracked over time. No schema drift is detected. The organization discovers the problem when a CRM field has been wrong for three months or a downstream pipeline quietly stops producing useful output.
\end{itemize}

We term this the \emph{memory governance gap}: the absence of an infrastructure layer governing what agents store, how stored information is typed and queried, which organizational policies reach which agent, how context is delivered across autonomous execution steps, and whether the system is performing reliably.

\subsection{Why RAG Is Not Enough}

RAG~\cite{lewis2020rag,gao2023ragsurvey} established a foundational paradigm: ground model outputs in retrieved evidence. But RAG is a \emph{retrieval primitive}, not an infrastructure layer. It addresses a single concern, retrieval relevance, and assumes a single agent, a single query, and a static document store. It provides no mechanism for governing what agents write into the store, no organizational context routing based on task requirements, no session-aware delivery across autonomous execution steps~\cite{liu2024lost}, no schema enforcement for downstream consumption, and no quality feedback loop for detecting degradation at scale. \sysname{} addresses the layer RAG leaves vacant.

\subsection{Contributions}

This paper makes four contributions, each addressing one or more of the five challenges above:

\begin{enumerate}[leftmargin=*]
  \item \textbf{A dual memory taxonomy with formal quality gates} \emph{(addresses memory silos and the downstream dead-end problem).} We distinguish \emph{open-set memory} (coreference-resolved atomic facts stored as vector embeddings) from \emph{schema-enforced memory} (typed property values governed by organizational schemas with confidence scores), processed in a single extraction pass with automated quality gates. The shared store enables any agent across the organization to read and write entity memory through a common interface.

  \item \textbf{Tiered governance routing with progressive context delivery} \emph{(addresses governance fragmentation and context redundancy).} A mechanism for selecting which organizational context should be injected into an agent's context window, supporting a fast governance-aware hybrid path (${\sim}850$ms average, E3) and a full two-stage LLM selection path (${\sim}2$--$5$s), with session-aware delta delivery that tracks previously injected context across autonomous multi-step executions.

  \item \textbf{Reflection-bounded retrieval with entity-scoped isolation} \emph{(addresses memory silos).} An iterative protocol checking evidence completeness and generating targeted follow-up queries within bounded rounds, combined with CRM-key-based entity scoping that enforces hard isolation across tenants and entities.

  \item \textbf{Schema lifecycle management with closed-loop self-evaluation} \emph{(addresses the downstream dead-end and silent quality degradation).} A lifecycle spanning AI-assisted schema authoring, interactive enhancement, criteria-based rubric scoring with execution logging, and automated per-property schema refinement.
\end{enumerate}

\section{Related Work}
\label{sec:related}

\subsection{Retrieval-Augmented Generation and Iterative Retrieval}

RAG~\cite{lewis2020rag} and its extensions, scale~\cite{borgeaud2022retro}, self-reflection~\cite{asai2024selfrag}, iterative correction~\cite{yan2024crag}, and structure-preserving hierarchical selection~\cite{wang2025bookrag,lumer2025rethinking,li2026deepread}, address retrieval quality but leave four gaps: no governance over what is stored, no organizational context routing, no session-aware delivery, and no quality feedback loop. Document expansion by query prediction~\cite{nogueira2019docexpansion} improves retrieval by appending LLM-generated queries to documents at index time; HyDE~\cite{gao2023hyde} takes the inverse approach, generating hypothetical documents for a given query. Our governance routing layer adapts the document-expansion strategy to organizational context (\Cref{sec:governance}), but the core contribution of this work is the governance architecture itself rather than novel retrieval primitives.

\subsection{Agent and Production Memory Systems}

Single-agent memory systems~\cite{park2023generative,shinn2023reflexion,yao2023react,packer2023memgpt,zhong2024memorybank,jha2025engram} model memory for individual agents in controlled environments; they do not address organizational context, schema enforcement, or multi-tenant isolation. Production memory layers, SimpleMem~\cite{liu2026simplemem} and Mem0~\cite{chhikara2025mem0}, formalize write-time atomization into coreference-resolved facts with semantic deduplication. These are memory \emph{primitives}: they address how individual facts are stored and retrieved. \sysname{} operates at the infrastructure layer above, extending any memory substrate with schema enforcement, organizational governance routing, entity-scoped multi-tenant isolation, and closed-loop quality feedback. The relationship is architectural, not competitive; the governed layer requires capable memory primitives underneath, and the contributions of this paper are orthogonal to improvements in atomic fact extraction.

\subsection{Evaluation and Context Delivery}

Domain-specific rubric scoring with execution trace capture extends LLM-as-judge~\cite{zheng2023judging} and reference-free evaluation~\cite{es2024ragas} to the governance setting. Liu et al.~\cite{liu2024lost} demonstrated that models under-use information in long-context middles, motivating session-aware progressive delivery that injects only delta content on each step.

\subsection{Positioning Summary}

Prior systems address individual memory capabilities in isolation: SimpleMem and Mem0 provide memory primitives (atomic fact extraction, deduplication); Self-RAG introduces reflective retrieval; MemGPT explores tiered memory management. These operate at the level of a single agent's memory. No prior system provides the shared infrastructure layer above, combining schema-enforced typed memory, organizational governance routing, progressive context delivery, closed-loop schema refinement, and multi-tenant entity isolation into a unified architecture accessible to any agent across an organization. \sysname{} addresses this gap.

\section{Architecture Overview}
\label{sec:architecture}

\begin{figure}[t]
  \centering
  \includegraphics[width=\linewidth]{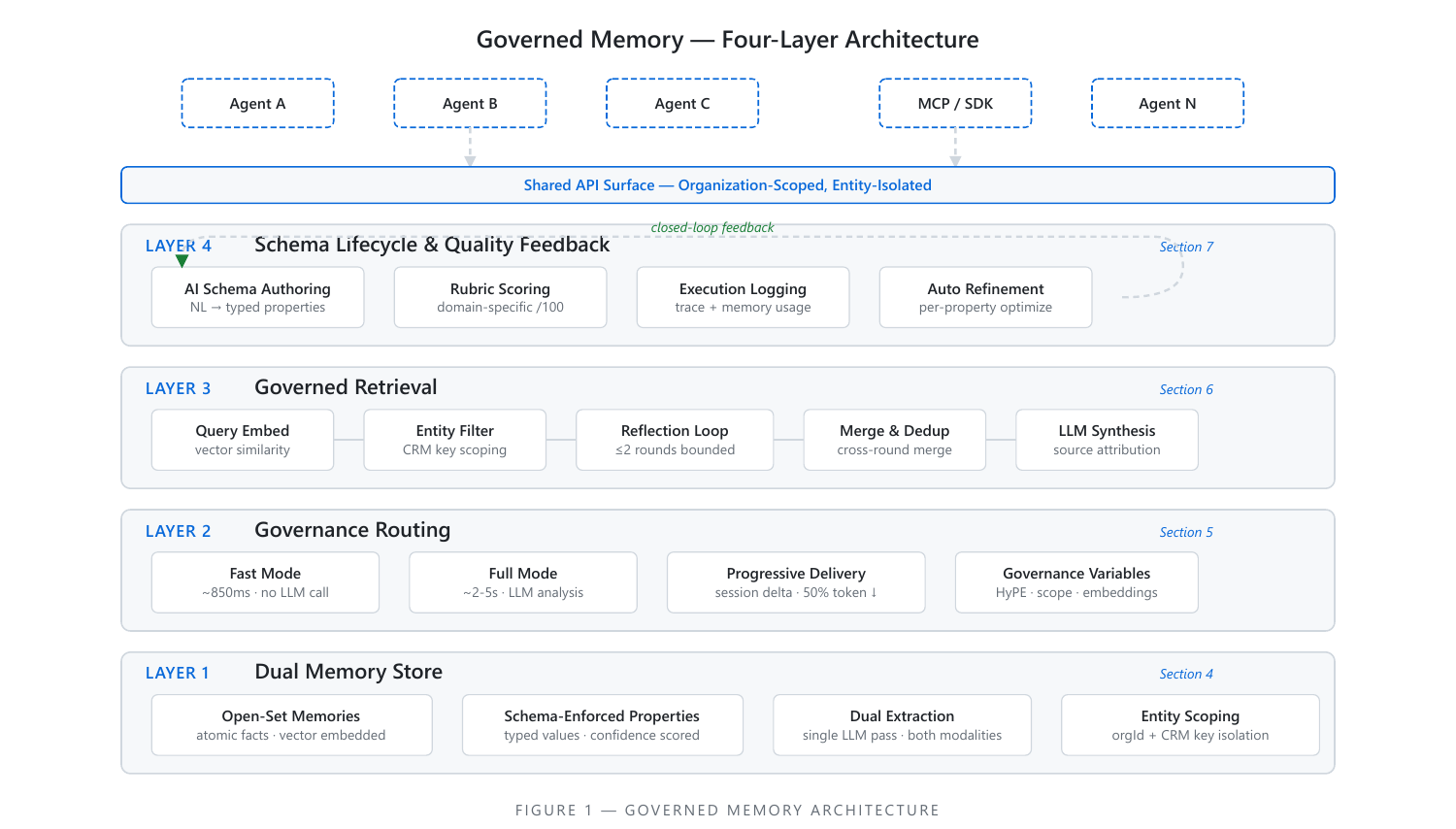}
  \caption{Governed Memory four-layer architecture. Agent nodes interact through a shared, organization-scoped API surface. Layers are independently configurable.}
  \label{fig:architecture}
\end{figure}

\sysname{} is organized as a four-layer architecture (\Cref{fig:architecture}). Each layer addresses a distinct governance concern and can be independently configured or disabled.

\textbf{Layer~1: Dual Memory Store.} Content is processed by a dual extraction pipeline producing \emph{open-set memories} (atomic, self-contained facts) and \emph{schema-enforced memories} (typed property values governed by organizational schemas), stored in a unified memory entry format with entity scoping and organizational isolation (\Cref{sec:dual-memory}).

\textbf{Layer~2: Governance Routing.} A tiered router selects which organizational context, policies, guidelines, templates, should be injected for a specific task. A \emph{fast} governance-aware hybrid path (${\sim}850$ms average, E3) ranks candidates without LLM calls; a \emph{full} two-stage path uses LLM structured analysis (${\sim}2$--$5$s). A session layer delivers only delta content (\Cref{sec:governance}).

\textbf{Layer~3: Governed Retrieval.} Vector similarity search with entity-scoped filtering and a reflection loop checking evidence completeness within bounded rounds (\Cref{sec:retrieval}).

\textbf{Layer~4: Schema Lifecycle and Quality Feedback.} AI-assisted schema authoring, domain-specific rubric evaluation with execution logging, and automated per-property refinement closing the feedback loop (\Cref{sec:schema-lifecycle}).

All memory entries share a unified record structure carrying entity scope, organizational partition, provenance metadata (content hash, extraction method, model identifier, chunk position, redaction status), and type-specific fields for schema-enforced properties. All operations are partitioned by organization ID, providing hard tenant isolation. Within an organization, retrieval can be further scoped to a specific entity using CRM keys, preventing cross-entity contamination. The system enforces a two-phase content redaction pipeline scrubbing PII and secrets before and after LLM extraction. Full data model, entity scoping, security posture, and redaction details are provided in \Cref{app:data-model,app:security}. The system exposes a standard MCP interface and SDK, enabling any compatible agent, regardless of framework or vendor, to read, write, and govern memory through the same organizational context without bespoke integration.

\section{Dual Memory Model}
\label{sec:dual-memory}

\subsection{Problem Statement}

Enterprise agents must store both unstructured insights and structured property values. A sales call transcript contains the free-form insight ``The CTO mentioned they are evaluating three vendors'' alongside the structured fact that the deal value is \$450,000. Existing memory systems support one modality or the other. The dual memory model stores both simultaneously from the same extraction pass, ensuring no information is lost to modality mismatch.

\subsection{Open-Set Memory}

\emph{Open-set memories} are atomic, self-contained facts extracted from unstructured content. The extraction prompt enforces five invariants: completeness, self-containment, coreference resolution, temporal anchoring, and atomicity. Three lightweight quality gates are computed per extraction batch, coreference score (pronoun detection), self-containment score (syntactic pattern matching), and temporal anchoring score (relative-time pattern detection), serving as early-warning operational signals. Open-set memories are embedded and stored in a vector database with per-organization partitioning. Before insertion, each candidate is compared against existing entries; candidates exceeding a cosine similarity threshold (default: 0.92) are skipped, preventing near-duplicate accumulation.

\subsection{Schema-Enforced Memory}

\emph{Schema-enforced memories} are typed property values extracted according to an organizational schema defining properties with names, descriptions, types (text, number, date, boolean, options, array), and extraction hints.

\textbf{Property selection.} Before extraction, the system selects relevant properties using embedding similarity between content and property metadata, with a minimum score threshold and maximum count cap. This prevents presenting the LLM with hundreds of irrelevant properties, reducing hallucination and improving type compliance.

\textbf{Dual extraction.} A single LLM call receives both content and selected property definitions, producing two parallel outputs: (1)~typed property values with confidence scores and update semantics, and (2)~open-set atomic facts. This ensures the same content is processed once. Extracted values are validated against schema-declared types. Values carry explicit update flags supporting both single-value replacement and temporal accumulation. Extraction confidence scores follow the observation that language models can calibrate their own certainty~\cite{kadavath2022calibration}.

\subsection{Algorithm: Dual Extraction Pipeline}

The extraction pipeline proceeds as: (1)~optional pre-extraction PII redaction; (2)~content chunking with overlap; (3)~per-chunk dual extraction with property selection; (4)~post-extraction redaction scan; (5)~cross-chunk deduplication (highest-confidence properties, normalized-text facts); (6)~quality gate computation and logging; (7)~embedding generation; (8)~provenance attachment; and (9)~write-side deduplication against the existing store. Algorithm description is provided in \Cref{app:algorithms} (Algorithm~2).

A background consolidation process (\Cref{app:consolidation}, Algorithm~3) periodically merges near-duplicate memories and prunes stale entries, using a deliberately higher similarity threshold (0.95) than write-side deduplication to minimize false merges.

\begin{figure}[t]
  \centering
  \includegraphics[width=\linewidth]{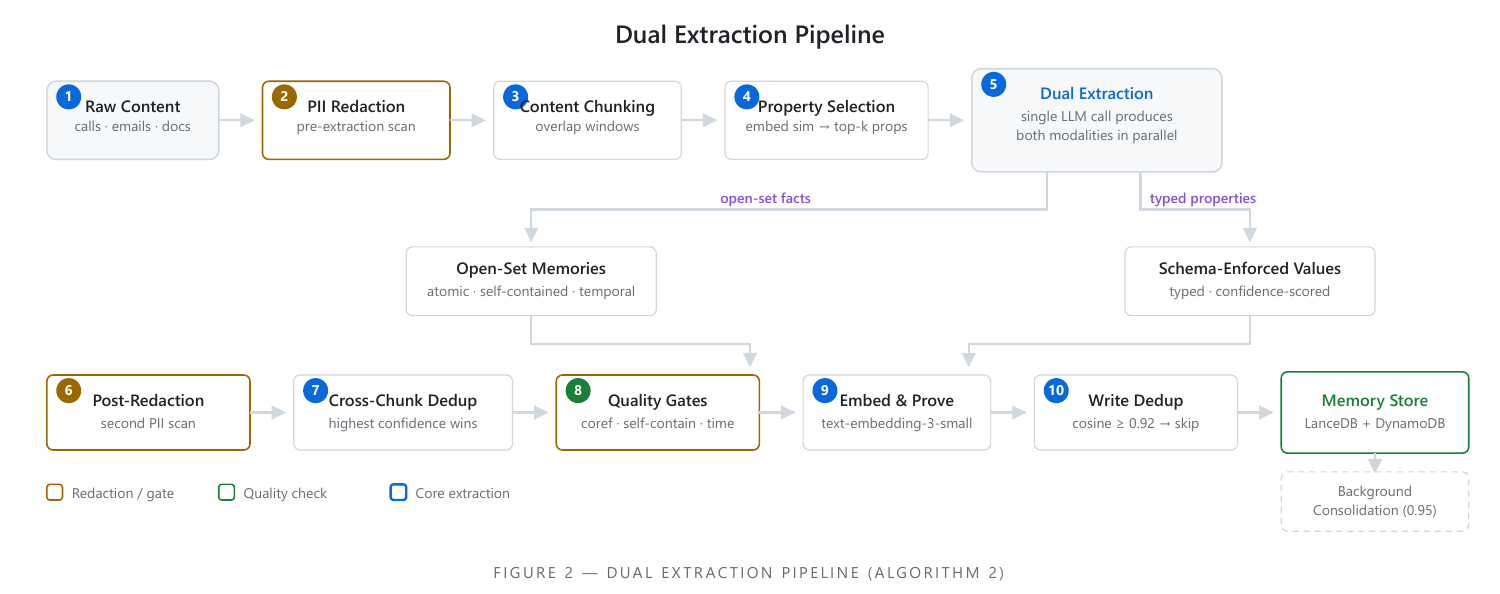}
  \caption{Dual extraction pipeline. A single LLM call produces both open-set facts and schema-enforced typed properties, followed by quality gates and write-side deduplication.}
  \label{fig:extraction}
\end{figure}

\section{Governance Routing}
\label{sec:governance}

\subsection{Governance Variable Model}

Organizational context is stored as \emph{governance variables}, also called guidelines with structured metadata including name, description, tags, content, heading hierarchy, and content-aware embeddings. When a variable is created or updated, three enrichment steps run in parallel: (1)~\emph{Hypothetical Prompt Enrichment (HyPE)}, generating synthetic queries representing plausible agent requests, adapting the document-expansion-by-query-prediction approach~\cite{nogueira2019docexpansion} (which appends predicted queries to documents before indexing) to the governance setting, with naming inspired by HyDE~\cite{gao2023hyde} (which instead generates hypothetical documents for a given query); (2)~\emph{governance scope inference}, an LLM determines whether the variable is always-on and infers trigger keywords; and (3)~\emph{content-aware embedding} computed from metadata and content preview.

\subsection{Tiered Routing Modes}

\textbf{Fast mode} (${\sim}850$ms average, E3). No LLM call. Each candidate is scored using a weighted composite of embedding similarity and keyword overlap against variable metadata and HyPE-generated queries, plus a governance scope boost for always-on variables. Results are partitioned into critical and supplementary sets with dynamic caps. Algorithm description is provided in \Cref{app:algorithms}.

\textbf{Full mode} (${\sim}2$--$5$s). Two-stage pipeline: embedding pre-filter reducing candidates, followed by LLM multi-step structured analysis classifying context as critical or supplementary with section-level extraction capability.

\textbf{Auto mode} (default). Selects fast or full based on library characteristics.

\subsection{Progressive Context Delivery}

Modern agents increasingly operate in autonomous multi-step loops, planning, acting, observing, re-planning, and acting again, without human intervention between steps. In such workflows, an agent may invoke governance routing multiple times within a single session as its task evolves and new context becomes relevant. Simply re-injecting the full governance set on every step creates three compounding problems: (1)~\emph{token bloat}, as context windows fill with previously delivered material rather than new task-relevant guidance; (2)~\emph{accuracy degradation}, as redundant context competes with fresh instructions for the model's attention~\cite{liu2024lost}; and (3)~\emph{unnecessary cost}, since each duplicated variable consumes tokens billed per step.

Progressive delivery addresses this by maintaining a session state record that tracks which variables (and which sections within them) have already been delivered. On each routing call, already-delivered variables are excluded; only new or newly relevant content is resolved and injected. Supplementary items are intentionally not recorded, allowing promotion on subsequent calls if the task evolves, ensuring that a guideline initially deemed peripheral can surface as critical when the agent's plan shifts. Algorithm description is provided in \Cref{app:algorithms} (Algorithm~5).

\begin{figure}[t]
  \centering
  \includegraphics[width=\linewidth]{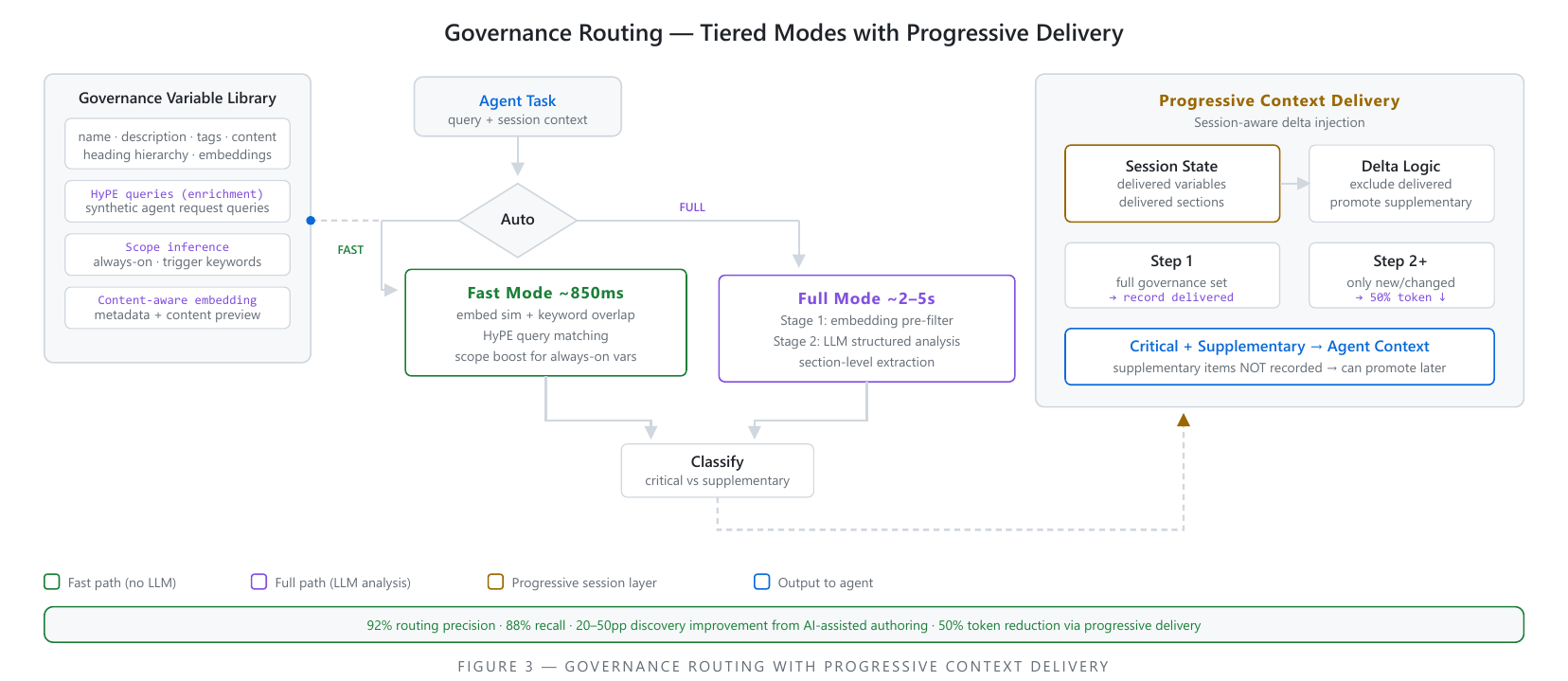}
  \caption{Governance routing with tiered modes (fast/full) and progressive context delivery. The session layer tracks delivered variables to inject only delta content on each autonomous step.}
  \label{fig:governance}
\end{figure}

\section{Reflection-Bounded Retrieval}
\label{sec:retrieval}

\subsection{Retrieval Architecture}

The retrieval path proceeds as: (1)~query embedding; (2)~vector search within the organization partition with entity-scoped CRM key filters; (3)~post-filtering by metadata (persons, entities, location, timestamp range, memory type); (4)~optional reflection loop; (5)~merge and deduplication; and (6)~optional LLM answer synthesis with source attribution.

\subsection{Reflection Loop}

The reflection loop is bounded by a configurable maximum round count (default: 2). Each round: an LLM judges evidence completeness at low temperature (0.1), and if incomplete, generates one to two targeted follow-up queries at moderate temperature (0.3). Results are merged by identifier across rounds. Each round adds predictable latency: one LLM call plus zero to two embedding-and-search operations. Algorithm description is provided in \Cref{app:algorithms} (Algorithm~6).

\subsection{Hybrid Retrieval}

Retrieval operates across both memory types simultaneously, returning results in a unified format. A standalone entity context injection endpoint compiles per-entity data from both storage tiers into a token-budgeted context block. The endpoint resolves entity identity through CRM keys, fetches schema-enforced property values and open-set memories, and compiles them into a structured block with Properties and Observations sections. Token budget enforcement prioritizes schema-enforced properties (more structured, actionable) over open-set memories (ordered by recency), enabling any downstream consumer to obtain entity context without invoking governance routing.

\begin{figure}[t]
  \centering
  \includegraphics[width=\linewidth]{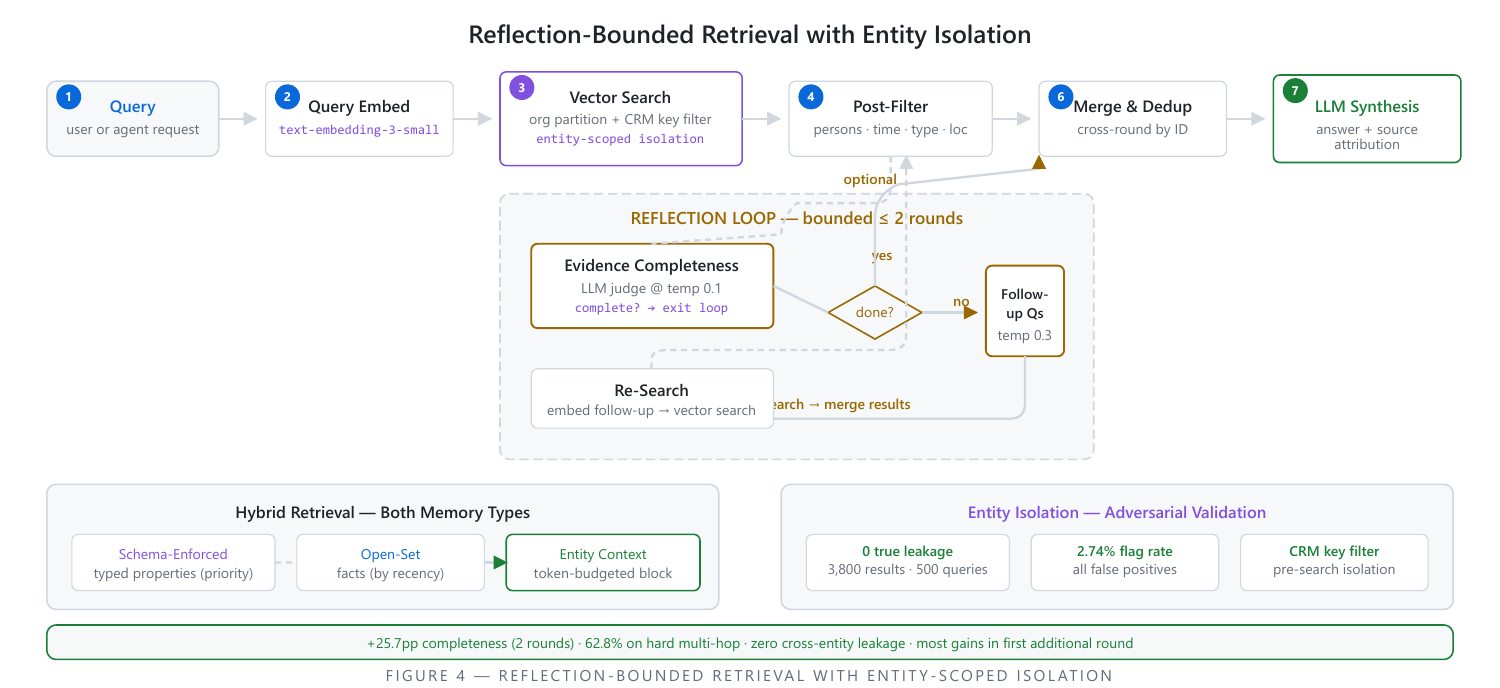}
  \caption{Reflection-bounded retrieval with entity-scoped isolation. The optional reflection loop generates targeted follow-up queries within bounded rounds; entity isolation is enforced by CRM key pre-filtering.}
  \label{fig:retrieval}
\end{figure}

\section{Schema Lifecycle and Self-Evaluation}
\label{sec:schema-lifecycle}

This section describes the full schema lifecycle, from creation through interactive improvement to automated refinement, designed around a central principle: schemas are living documents that evolve through human guidance and AI-augmented feedback.

\subsection{Schema Authoring and Interactive Enhancement}

\textbf{AI-assisted authoring.} Users describe, in natural language, what structured information they need to extract from their data, e.g., ``I want to capture each contact's role, buying intent, and preferred communication channel.'' An AI assistant within the web application translates this description into a complete schema definition, generating typed property specifications with names, descriptions, extraction instructions, and validation constraints. This streaming generation service accelerates the translation from user intent to well-formed schema without requiring expertise in schema design.

\textbf{Interactive enhancement.} Once a schema is in use, operators describe observed issues in natural language, e.g., ``the role field is too vague'' or ``buying intent should distinguish between active evaluation and future interest'', and receive streaming revised property definitions, at per-property or bulk granularity.

\subsection{Criteria-Based Rubric Scoring}

\Cref{sec:schema-lifecycle-scoring,sec:schema-lifecycle-logging,sec:schema-lifecycle-aggregation} are currently available via API to our internal team and select early-access users for iterative accuracy and performance improvement.
\label{sec:schema-lifecycle-scoring}

Users can invoke the platform's native agents or prompt endpoints via API, and then evaluate how governance context and retrieved memory influenced output quality. Each agent interaction is scored against domain-specific rubrics. The system provides four presets (default, sales, support, research), each normalized to 100 points with weighted criteria. Organizations can define custom rubrics. Evaluation bias is mitigated through rubric-first prompting, trace-grounded evaluation (access to full execution trace including tool calls and memory operations), and configurable cross-model evaluation.

\subsection{Execution Logging}
\label{sec:schema-lifecycle-logging}

Beyond scoring, evaluation captures a structured trace: conversation summary, tool usage log, memory recall log (with usage flags), memory creation log, and governance context log with helpfulness ratings. This enables diagnostic questions scalar scores cannot answer: ``Did the agent score low on Completeness because it failed to recall relevant memories, or because it recalled them but did not use them?''

\subsection{Aggregation and Quality Feedback}
\label{sec:schema-lifecycle-aggregation}

Evaluation records enable low-score detection, per-criterion breakdown, trend analysis annotated with schema and model changes, and per-endpoint comparison across models. Low context utilization scores inform governance improvements; low recall completeness scores inform reflection configuration. Recurring diagnostic patterns are summarized in \Cref{app:schema-example}, \Cref{tab:diagnostic-patterns}.

\subsection{Automated Schema Refinement}

Schema refinement is available to all users through the web application. When extraction quality is evaluated, the system executes a three-phase pipeline: (1)~\emph{extraction replay} producing baseline results; (2)~\emph{per-property analysis} classifying each property as extracted, missed, low-confidence, inaccurate, or unavailable, generating structured improvement instructions; (3)~\emph{parallel per-property optimization} producing revised definitions with change annotations. The three-phase design separates objective data, diagnostic judgment, and targeted fixes. Running Phase~3 in parallel keeps latency bounded. Algorithm description is provided in \Cref{app:algorithms} (Algorithm~7) and a worked example in \Cref{app:schema-example}.

\section{Experimental Evaluation}
\label{sec:experiments}

\subsection{Evaluation as Operational Monitoring}

These experiments serve a dual purpose: validating the core architectural mechanisms and defining a \emph{repeatable evaluation methodology} that organizations can apply as ongoing operational monitoring, running continuously as schemas, content types, and underlying models evolve. Each experiment targets a specific governance concern and maps directly to a production health signal: extraction quality, routing precision, retrieval completeness, entity isolation, conflict detection, and end-to-end output quality. The metrics introduced here, governance routing precision and recall, schema discovery rate, context defect rate, and memory density curves, are proposed as standard instrumentation for production memory systems, not one-time benchmark scores. Small controlled datasets are intentional: they are monitoring templates, designed for fast, interpretable re-execution, not underpowered one-shot studies. Sample sizes were validated by scaling: experiments began at $N{=}10$ per content type and were incrementally increased to $N{=}50$; key metrics (fact recall, routing precision, entity isolation) stabilized with minimal variance beyond $N{=}30$ per type, confirming that $N{=}250$ is sufficient for the effects being measured.

We designed experiments using synthetic datasets with embedded ground truth, enabling reproducible measurement with known fact counts, property values, coreference issues, and near-duplicates. All experiments use a fixed random seed (42) and were executed against the production API. Synthetic data results are designed to stress distinct extraction challenges under controlled conditions.

\begin{table}[t]
\centering
\caption{Experimental dataset summary.}
\label{tab:datasets}
\small
\begin{tabular}{@{}lll@{}}
\toprule
\textbf{Dataset} & \textbf{Samples} & \textbf{Ground Truth Elements} \\
\midrule
Primary corpus & 250 samples, 5 types & 8--12 facts, 5--8 props/sample \\
Multi-source entity & 5 sources & 40 unique facts, 8 cross-source dupes \\
Entity isolation profiles & 100 entities & High overlap with unique markers \\
Recall query sets & 500 queries & Expected topics \& min source counts \\
Governance variable pairs & 5 pairs & Known targets for 15 tasks \\
Conflict pairs & 30 pairs, 15 cats & Stale + fresh claims with known dates \\
\bottomrule
\end{tabular}
\end{table}

\subsection{Extraction Quality Across Content Types (E1)}

\begin{table}[t]
\centering
\caption{Extraction quality by content type (250 samples).}
\label{tab:extraction}
\small
\begin{tabular}{@{}lcc@{}}
\toprule
\textbf{Content Type} & \textbf{Samples} & \textbf{Fact Recall} \\
\midrule
Call notes & 50 & \textbf{100\%} \\
Documents & 50 & \textbf{100\%} \\
Emails & 50 & \textbf{100\%} \\
Transcripts & 50 & \textbf{100\%} \\
Chats & 50 & 98\% \\
\midrule
\textbf{Overall} & \textbf{250} & \textbf{99.6\%} \\
\bottomrule
\end{tabular}
\end{table}

Fact recall is consistently 99--100\% regardless of content format. We acknowledge that these near-perfect scores reflect evaluation on synthetic datasets that, while structurally diverse across five content types and designed to stress distinct extraction challenges (coreference, temporal reasoning, implicit facts), are free of the noise, formatting inconsistencies, and ambiguity typical of production data. The full evaluation datasets and ground-truth annotations are publicly available at \url{https://github.com/personizeai/governed-memory}. The results should therefore be interpreted as evidence that the extraction architecture and algorithm reliably capture ground-truth facts under controlled, diverse conditions, an upper-bound demonstration of the pipeline's capability, rather than as a claim of identical performance on arbitrary real-world inputs. Production deployments exhibit comparable but modestly lower recall, consistent with the additional noise in organic content. The schema refinement mechanisms (\Cref{sec:schema-lifecycle}) address property extraction variance, which is predominantly attributable to content structure and schema maturity rather than algorithmic limitations.

\subsection{Quality Gates Ablation (E9)}

On 40 samples, retrieval with quality gates reduces the output defect rate by 25\% relative compared to raw retrieval (6.3\% vs.\ 8.4\%). Temporal accuracy improves by 6.8 percentage points (95.2\% vs.\ 88.4\%), and signal-to-noise ratio (useful facts retrieved vs.\ noise) increases from 1.1:1 to 4.2:1. These gains arise from the write-time quality gate pipeline filtering coreference-unresolved, non-self-contained, and temporally ambiguous facts before they enter the store, downstream retrieval inherits cleaner signal. Decision precision (94.5\%) validates the gate as a reliable guard rather than a coarse filter.

\subsection{Dual Memory Complementarity (E12)}

\begin{table}[t]
\centering
\caption{Coverage distribution across memory modalities (20 samples).}
\label{tab:complementarity}
\small
\begin{tabular}{@{}lc@{}}
\toprule
\textbf{Category} & \textbf{\% of Total} \\
\midrule
Captured by both modalities & 34\% \\
Open-set only (long-tail insights) & 38\% \\
Schema-enforced only (typed values) & 12\% \\
Missed by both & 16\% \\
\midrule
\textbf{Combined recall} & \textbf{82.8\%} \\
\bottomrule
\end{tabular}
\end{table}

The 38\% captured exclusively by open-set memory, relational facts, qualitative observations, contextual details, would be permanently lost in a schema-only system. The 12\% captured exclusively by schema enforcement would lack type enforcement in an open-set-only system. This validates the dual architecture as complementary rather than redundant.

Two unique advantages of schema-enforced properties are worth noting. First, the schema is not fixed: the schema refinement mechanisms (\Cref{sec:schema-lifecycle}) allow operators to evolve property definitions over time, progressively promoting observed patterns into typed, queryable structure as the deployment matures. Second, and more consequential for production use: schema-enforced properties fulfill a role that open-set memories, which are stored as free-form text, cannot. Structured properties are directly addressable by downstream query logic, filtering pipelines, and expert systems; they carry type guarantees and can drive conditional routing and decision logic without natural language parsing. A compliance constraint or communication preference that exists only as free-form memory is retrievable but not directly queryable; the same fact captured as a schema-enforced property becomes a first-class signal in any downstream system. The practical ceiling of the dual architecture is therefore higher than the coverage numbers alone suggest: the 38\% open-set-only bucket preserves recall that would otherwise be lost, while the schema layer, continuously refinable, determines how much of that captured knowledge becomes structurally actionable.

\subsection{Governance Routing Effectiveness (E3, E13)}

\textbf{Routing precision.} Against 25 governance variables across 5 categories, routing achieves \textbf{92\% precision} (nearly all selected variables are relevant) and \textbf{88\% recall} across 20 diverse task types spanning sales, compliance, engineering, marketing, and support.

\textbf{Authoring quality impact.} Well-authored governance variables are \textbf{20--50 percentage points more discoverable} than poorly-authored equivalents: in 3 of 5 categories (brand, product, support), poorly-authored variables scored 0\% discovery rate. This validates the AI-assisted authoring tools (\Cref{sec:governance}, \Cref{sec:schema-lifecycle}) as operationally significant.

\subsection{Reflection-Bounded Retrieval (E10)}

\begin{table}[t]
\centering
\caption{Reflection ablation on 10 hard multi-hop queries (4 conditions).}
\label{tab:reflection}
\small
\begin{tabular}{@{}lccc@{}}
\toprule
\textbf{Condition} & \textbf{Avg Compl.} & \textbf{Avg Results} & \textbf{Avg Latency} \\
\midrule
No reflection (baseline) & 37.1\% & 15.0 & 9.4s \\
API-managed, 1 round & 40.4\% & 22.5 & 10.4s \\
Manual multi-hop, 1 round & 61.2\% & 20.8 & 6.5s \\
\textbf{Manual multi-hop, 2 rounds} & \textbf{62.8\%} & 21.9 & 10.0s \\
\bottomrule
\end{tabular}
\end{table}

Manual multi-hop retrieval with 2 rounds achieves \textbf{62.8\% completeness versus 37.1\% baseline}, a 25.7 percentage point improvement on hard multi-faceted queries where information is scattered across 3--5 sources. API-managed reflection (+3.3pp) shows more modest gains than manual multi-hop (+25.7pp), indicating that query generation strategy is the key determinant of reflection effectiveness. This gap does not reflect a retrieval quality difference between the two paths; it reflects the investment in query strategy. API-managed reflection applies generic follow-up queries; manual multi-hop allows developers to decompose complex questions into targeted retrieval passes tuned to their domain. The completeness ceiling on hard multi-faceted queries is therefore set by application-layer query design, not by the memory system itself. The single-round manual condition (61.2\%) approaches the two-round ceiling, suggesting most completeness gains materialize in the first additional retrieval pass. Reflection is effective when the memory store contains relevant but scattered information; cases where underlying data is absent yield diminishing returns regardless of round count.

\subsection{Entity Isolation (E11)}

Under adversarial conditions, 100 entities with same industry, similar roles, overlapping names, and similar deal sizes, entity-scoped retrieval produces \textbf{zero true cross-entity leakage} across 3,800 results (500 queries $\times$ 5 query types). Of the 2.74\% observed flag rate (104 flags), all are false positives attributable to shared name tokens across distinct records, not actual memory bleed. Isolation is enforced by the CRM key pre-filtering mechanism, not embedding distinctiveness.

\subsection{Semantic Conflict Resolution (E14)}

When the same entity accumulates contradictory facts over time (e.g., a company changes its primary database), the system must surface the most recent claim. E14 tests this with 30 conflict pairs across 15 categories (database, cloud provider, team size, budget, etc.), each consisting of a stale memorization (74--270 days old) followed by a fresh memorization (0--57 days old) for the same contact. Retrieval applies exponential recency decay (half-life $= 38$ days) to rank recent facts above outdated ones.

The primary correctness measure is conflict detection: whether the fresh claim is present in the answer at all. Full stale suppression applies a stricter standard: the answer must contain only the fresh claim with zero reference to stale content.

\begin{table}[t]
\centering
\caption{Semantic conflict resolution (30 pairs).}
\label{tab:conflict}
\small
\begin{tabular}{@{}p{3.2cm}p{6.5cm}c@{}}
\toprule
\textbf{Metric} & \textbf{Definition} & \textbf{Result} \\
\midrule
Conflict detection \emph{(primary)} & Fresh info surfaced in answer & \textbf{83.3\%} \\
Full stale suppression \emph{(strict)} & Answer contains \emph{only} fresh keywords & 33.3\% \\
Incorrect stale & Answer reflects only the outdated claim & 3.3\% \\
\bottomrule
\end{tabular}
\end{table}

Of the 15 \emph{both\_present} verdicts, all 15 answers lead with the fresh claim and reference the stale value only as transition context (e.g., ``migrated from AWS to Google Cloud''). Effective answer correctness, fresh claim present and presented as current, is \textbf{83.3\%}. The single incorrect-stale case involved a pain-point category where the fresh and stale claims shared overlapping vocabulary, causing keyword-based evaluation ambiguity.

\subsection{End-to-End Ablation (E8)}

\begin{table}[t]
\centering
\caption{End-to-end evaluation (10 prospects, 3 runs each, sales rubric /100).}
\label{tab:e2e}
\small
\begin{tabular}{@{}lcc@{}}
\toprule
\textbf{Condition} & \textbf{Avg Score /100} & \textbf{$\Delta$ vs.\ Baseline} \\
\midrule
A: No memory & 79.5 & , \\
B: Raw memory & 85.2 & +5.7 \\
C: Open-set + governance & 86.4 & +6.9 \\
\textbf{D: Full governed memory} & \textbf{85.9} & \textbf{+6.4} \\
\bottomrule
\end{tabular}
\end{table}

\textbf{Memory provides the primary quality gain} (A$\to$B: +5.7 pts). Governance routing adds measurable refinement (B$\to$C: +1.2 pts). Full governed memory (D, +6.4 pts) scores comparably to C on this rubric ($-0.5$ pts), an expected result given that email-generation quality metrics measure tone, framing, and personalization, dimensions where open-set facts and governance context already provide the dominant signal. Schema enforcement's primary value is realized downstream of generation: typed, validated property values enable reliable CRM synchronization, analytics aggregation, and structured API consumption that are orthogonal to single-interaction quality scores. Organizations can iteratively improve schema extraction accuracy using the AI-assisted schema optimization tools (\Cref{sec:schema-lifecycle}), refining property definitions based on observed extraction patterns and measuring downstream impact through the same self-evaluation pipeline. The combined system (D, +6.4 pts) consistently outperforms the no-memory baseline across all runs, validating the core thesis. It should be noted that the rubric used here is likely an underestimate of the true quality differential: a rubric optimized for single-interaction email quality cannot capture the compounding gains that schema enforcement and governed memory produce at scale, across repeated interactions, multi-record aggregation, conditional personalization logic, and downstream system reliability. A rubric designed to measure those dimensions would be expected to show a larger separation between conditions.

\subsection{Adversarial Governance (E15)}

Against 50 adversarial scenarios designed to bypass governance constraints, distributed across easy, medium, and hard difficulty, the system achieves \textbf{100\% compliance} across all difficulty levels and a \textbf{96\% guardrail activation rate} (48/50 scenarios triggered the intended guardrail; 2 easy-category inputs resolved correctly without explicit guardrail invocation). Zero organizational policy leakage was observed. This validates the governance layer as robust to deliberate constraint circumvention attempts.

\subsection{External Benchmark Validation: LoCoMo}

As external validation that the system functions as a general-purpose long-term memory system, we evaluated on \textbf{LoCoMo}~\cite{maharana2024locomo} (272 sessions, 1,542 questions across 10 conversations).

\begin{table}[t]
\centering
\caption{LoCoMo benchmark results.}
\label{tab:locomo}
\small
\begin{tabular}{@{}lcc@{}}
\toprule
\textbf{Category} & \textbf{Accuracy} & \textbf{vs.\ Human Baseline} \\
\midrule
Single-hop & \textbf{78.7\%} & $-$16.4pp \\
Multi-hop & 51.7\% & $-$34.1pp \\
Temporal & 64.6\% & $-$28.0pp \\
Open-ended & \textbf{83.6\%} & \textbf{+8.2pp} \\
\midrule
\textbf{Overall} & \textbf{74.8\%} & $-$13.1pp \\
\bottomrule
\end{tabular}
\end{table}

The system achieves 74.8\% overall against a human baseline of 87.9\%, \textbf{exceeding human-level on open-ended inference} (83.6\% vs.\ 75.4\%), the largest category (841 questions). This result confirms that the governance, schema enforcement, and entity isolation layers impose no retrieval quality penalty, the architecture achieves state-of-the-art memory accuracy \emph{despite} doing substantially more than standalone memory systems. For context, independently evaluated memory systems report 42--67\% on comparable settings: Mem0~\cite{chhikara2025mem0} at 64--67\%, Zep at 42--66\%, OpenAI built-in memory at ${\sim}53$\%. Multi-hop (51.7\%) and temporal (64.6\%) remain active optimization areas. Results use a hybrid text-match-first / LLM-judge-fallback methodology; 950 of 1,153 correct answers (82.4\%) scored via text-match. Published systems use varying methodologies, pure token-overlap F1 or pure LLM-as-judge, producing scores not directly comparable across systems.

\section{Discussion}
\label{sec:discussion}

\subsection{Limitations}

\textbf{Quality gates are heuristic.} Coreference, self-containment, and temporal anchoring scores use pattern-based heuristics rather than deep semantic analysis. Calibration against human judgments would strengthen confidence. \textbf{Self-evaluation relies on self-judgment}, with mitigations (rubric-first prompting, cross-model evaluation) that reduce but do not eliminate LLM-as-judge biases~\cite{zheng2023judging}. \textbf{Deduplication thresholds are empirically tuned} (0.92 write-side, 0.95 consolidation); adaptive thresholds based on content characteristics would be more robust. \textbf{Redaction is regex-based}~\cite{microsoft2024presidio}, covering well-structured PII with high precision but potentially missing obfuscated or context-dependent patterns. \textbf{Multi-agent write conflicts are unvalidated.} E14 evaluates temporal conflicts between sequential writes (stale vs.\ fresh claims over time) but does not test concurrent writes from multiple agents acting on the same entity simultaneously. As a shared layer serving many agents, concurrent write conflicts are a realistic production scenario; conflict detection and resolution under concurrent conditions remain an open problem.

\subsection{Design Tensions}

Open-set vs.\ schema-enforced memory is a spectrum: some facts extracted as open-set could map to schema properties. The dual model accepts this overlap deliberately, ensuring no information is lost. Reflection yields strong gains (+25.7pp) under optimal query generation but shows diminishing returns when underlying data is absent; the gap between manual multi-hop (62.8\%) and API-managed reflection (40.4\%) indicates that query generation strategy, rather than round count, is the primary lever (\Cref{sec:experiments}). Making the round bound query-adaptive and improving API query generation are planned future work. Progressive delivery assumes previously delivered context remains relevant; the 24-hour session TTL limits but does not eliminate staleness.

\subsection{Future Work}

Key remaining gaps include: \textbf{cross-organization validation} measuring whether patterns hold across varying content and schema maturity, including multi-agent write compatibility and contradiction detection; \textbf{automated schema expansion} proposing new properties from unextracted content patterns; \textbf{hybrid retrieval} combining semantic search with keyword lanes and reciprocal rank fusion; and \textbf{ML-augmented redaction} supplementing regex patterns with transformer-based NER for non-standard PII.

\subsection{Ethical Considerations}

The system stores extracted facts about people, and memory accumulation creates detailed profiles. Organizations bear responsibility for regulatory compliance (GDPR, CCPA) and data subject rights. Schema-enforced extraction reflects biases in both language models and operator-defined schemas; organizations should review schemas for discriminatory patterns. The governance layer provides a mitigation: organizational policies are explicit, auditable artifacts rather than implicit model behaviors.

\section{Conclusion}
\label{sec:conclusion}

This paper identified the \emph{memory governance gap}, five structural challenges arising when dozens of autonomous agent nodes act on the same entities across workflows without shared memory or common governance, and presented \sysname{}, a shared memory and governance layer addressing it through four integrated mechanisms. Experimental validation confirms that each mechanism works at production scale: 99.6\% fact recall with complementary dual-modality coverage; 92\% governance routing precision; zero cross-entity leakage across 3,800 adversarial queries; 100\% adversarial governance compliance; and 74.8\% on the LoCoMo benchmark, confirming that governance, schema enforcement, and entity isolation impose no retrieval quality penalty. Output quality saturates at approximately seven governed memories per entity, establishing a practical operating point for agentic deployments.

The terminology introduced, governed memory, governance routing, progressive context delivery, memory quality gates, schema lifecycle management, provides reference points for the growing community building production memory systems for agentic workflows. As autonomous agent deployments scale across organizations, accuracy and compliance across distributed agent nodes become architectural requirements, not afterthoughts. The system described is commercially deployed; code, datasets, and extended supplementary material are available at \url{https://github.com/personizeai/governed-memory}.
\newpage
\bibliographystyle{plainnat}

\newpage
\appendix

\section{Data Model and Entity Scoping}
\label{app:data-model}

All memory entries share a unified record structure:

\begin{lstlisting}
MemoryEntry:
  id              : string       -- unique identifier
  text            : string       -- atomic fact or normalized property statement
  orgId           : string       -- organizational partition key
  recordId        : string?      -- entity scope (contact, company, deal)
  type            : string       -- "memory" | "property_value"
  keywords        : string[]     -- extracted keywords
  persons         : string[]     -- mentioned persons
  entities        : string[]     -- mentioned entities
  location        : string?      -- geographic reference
  topic           : string?      -- topic classification
  timestamp       : string?      -- temporal anchor (ISO 8601)
  customAttributes: map          -- schema-enforced typed values
  source          : string       -- provenance marker
  score           : float?       -- relevance score (populated during retrieval)
  createdAt       : string
  updatedAt       : string

  -- Property-specific fields (when type = "property_value")
  propertyId      : string?      -- schema property identifier
  propertyName    : string?      -- human-readable property name
  systemName      : string?      -- canonical system name
  propertyValue   : string?      -- extracted value (serialized)
  collectionId    : string?      -- schema collection scope
  confidence      : float?       -- extraction confidence (0.0--1.0)

  -- Provenance metadata (stored in customAttributes)
  provenance      : map?
    contentHash      : string    -- SHA-256 of first 1000 chars
    contentLength    : int       -- original source content length
    speaker          : string?   -- identified speaker
    extractionMethod : string    -- "single_extract" | "dual_extract"
    llmModel         : string?   -- model identifier
    chunkIndex       : int?      -- chunk position (0-based)
    chunkTotal       : int?      -- total chunks in source
    redactionApplied : boolean?  -- whether redaction was applied
    timestamp        : string    -- ISO 8601 extraction timestamp
\end{lstlisting}

The \texttt{contentHash} enables tracing entries back to source documents. When content is chunked, \texttt{chunkIndex} and \texttt{chunkTotal} record position for reconstruction.

\textbf{Entity scoping.} All operations are partitioned by \texttt{orgId}. Within an organization, retrieval is scoped to entities using CRM keys:

\begin{lstlisting}
CRMKeys:
  recordId          : string?
  email             : string?
  websiteUrl        : string?
  phoneNumber       : string?
  customIdentifiers : map?
\end{lstlisting}

Entity types are open-ended: contacts, companies, deals, vendors, partners, devices, locations, and content assets share the same mechanisms.

\section{Security, Privacy, and Content Redaction}
\label{app:security}

\textbf{Security posture.} Data is encrypted at rest and in transit. Organization-level partition keys provide hard tenant isolation at the storage layer. Governance variable visibility (\texttt{organization}, \texttt{private}, \texttt{adminsOnly}) and access levels (\texttt{readOnly}, \texttt{cloneable}, \texttt{editable}) are enforced at the API layer. All operations are logged with timestamps, user identifiers, and operation metadata.

\textbf{Two-phase content redaction.} A redaction pipeline scrubs PII and secrets before and after LLM extraction:

\textbf{Phase~1 (Pre-Extraction).} Raw text is scanned for sensitive patterns; matches are replaced with typed placeholders, ensuring the LLM never sees original values.

\textbf{Phase~2 (Post-Extraction).} Extracted values are scanned again, catching cases where the LLM reconstructs PII-like patterns from contextual cues.

Entity detection is organized into four tiers:

\begin{table}[H]
\centering
\small
\begin{tabular}{@{}clll@{}}
\toprule
\textbf{Tier} & \textbf{Category} & \textbf{Entity Types} & \textbf{Detection} \\
\midrule
1 & Secrets & API keys, private keys, passwords & Pattern matching \\
2 & Financial PII & Credit cards, IBAN & Regex + Luhn \\
3 & Identity PII & Social Security Numbers & Regex + validation \\
4 & Contact PII & Emails, phones, IPs & Format matching \\
\bottomrule
\end{tabular}
\end{table}

Three anonymization strategies are supported: redact (typed placeholder), mask (preserve last 4 digits), hash (SHA-256 prefix for linkability without reversibility).

\begin{lstlisting}
Algorithm 1: Content Redaction

Input:  text (string), config (RedactionConfig)
Output: redactedText (string), audits (RedactionAudit[])

1. for each entityPattern in ENTITY_PATTERNS:
2.     if entityPattern.tier not enabled in config: continue
3.     if entityPattern is EMAIL and config.skipEmails: continue
4.     if entityPattern is PHONE and config.skipPhones: continue
5.     matches <- findAll(text, entityPattern.regex)
6.     for each match in matches:
7.         if entityPattern.validate and not validate(match):
8.             continue
9.         text <- replace(text, match,
                           applyStrategy(match, config.strategy))
10.        count += 1
11.    if count > 0: audits.append({tier, entityType, count})
12. return {text, audits}
\end{lstlisting}

\section{Algorithm Descriptions}
\label{app:algorithms}

This appendix summarizes the core algorithms referenced throughout the paper. Detailed algorithmic specifications, experiment protocols, and synthetic datasets are publicly available at \url{https://github.com/personizeai/governed-memory}.

\textbf{Algorithm~2: Dual Extraction Pipeline.} The pipeline proceeds in nine stages: (1)~optional pre-extraction PII redaction; (2)~content chunking with content-type-specific overlap parameters (dialogue, transcript, and document modes); (3)~per-chunk embedding-based property selection from the organizational schema; (4)~per-chunk dual LLM extraction producing both open-set atomic facts and typed property values in a single call; (5)~post-extraction redaction scan; (6)~cross-chunk deduplication prioritizing highest-confidence properties and normalized-text deduplication for facts; (7)~quality gate computation (coreference, self-containment, temporal anchoring); (8)~batch embedding generation with provenance attachment; and (9)~write-side deduplication against the existing store using cosine similarity thresholds.

\textbf{Algorithm~3: Background Consolidation.} A two-phase process: (1)~merge near-duplicate memories using a deliberately higher similarity threshold than write-side deduplication to minimize false merges; (2)~prune stale memories beyond a configurable retention window, with sole-memory protection ensuring no entity is left with zero open-set memories.

\textbf{Algorithm~4: Embedding Pre-Filter.} Reduces the governance variable candidate set before LLM-based routing. Candidates without embeddings pass through unconditionally. Remaining candidates are scored by cosine similarity against the task embedding; those exceeding a minimum score threshold or ranking in the top-$K$ are retained.

\textbf{Algorithm~5: Tiered Governance Routing.} Two paths depending on resolved mode. \emph{Fast path:} the task message is embedded and scored against all candidates using a composite of embedding similarity and keyword overlap against variable metadata and synthetic queries; always-on variables receive unconditional inclusion; results are partitioned into critical and supplementary sets; session state is consulted to exclude already-delivered variables before returning compiled context. \emph{Full path:} an embedding pre-filter reduces the candidate set, then an LLM performs multi-step structured analysis classifying each variable as critical or supplementary with section-level precision; fallback promotion applies if no critical selections result. Both paths return a compiled critical context block and supplementary metadata.

\textbf{Algorithm~6: Reflection-Bounded Retrieval.} An iterative loop bounded by a configurable maximum round count. Each round: (1)~an LLM judges evidence completeness against the original query at low temperature; (2)~if incomplete, the LLM generates targeted follow-up queries at moderate temperature; (3)~follow-up queries are embedded and used for additional vector searches; (4)~results are merged by identifier across rounds. The loop terminates when the completeness check passes or no further follow-up queries are generated.

\textbf{Algorithm~7: Automated Schema Refinement Pipeline.} A three-phase process: (1)~extraction replay producing baseline results; (2)~per-property analysis classifying each property as extracted, missed, low-confidence, inaccurate, or unavailable, with structured improvement instructions; (3)~parallel per-property optimization producing revised definitions with change annotations. The three-phase design separates objective data, diagnostic judgment, and targeted fixes.

\section{Background Consolidation Details}
\label{app:consolidation}

\textbf{Merge threshold selection.} The merge threshold (0.95 cosine similarity) is set higher than write-side deduplication (0.92). At write time, the system errs toward preventing duplicates; during consolidation, the system errs toward preserving distinct memories, since false merge cost exceeds near-duplicate retention cost.

\textbf{Sole-memory protection.} The prune phase ensures no entity is left with zero open-set memories. If all memories for an entity are older than the retention cutoff, the most recent is preserved.

\textbf{Operational design.} Organizations below a minimum memory count (default: 10) are skipped. A dry-run mode logs merge/prune decisions without executing deletions. An optional compaction step triggers storage compaction one hour after consolidation.

\section{Governance Routing Details}
\label{app:governance-details}

\subsection*{Stage~2: LLM Multi-Step Structured Selection}

The LLM performs a four-step structured analysis: (1)~\emph{Task Understanding}, restating the objective and implicit requirements; (2)~\emph{Quality Dimension Identification}, determining which dimensions matter (tone, compliance, structure, etc.); (3)~\emph{Task Refinement}, rewriting the task as a precise instruction; (4)~\emph{Selection and Prioritization}, for each variable, specifying priority (critical/supplementary), mode (full/section), and reasoning.

\textbf{Fallback promotion.} If no critical selections are returned but supplementary exist, the top two supplementary items are promoted.

\subsection*{Section-Level Extraction}

When mode ``section'' is selected, only requested sections are extracted using heading hierarchy boundaries. If a heading is not found, full content is delivered as fallback.

\subsection*{AI-Assisted Governance Authoring}

\textbf{Generation from intent.} The system generates governance content from variable name and description alone, producing actionable, structured content (criteria with bullet points, guidelines with sections, templates with placeholders, procedures with steps).

\textbf{Iterative refinement.} Users refine content through natural-language feedback, preserving format while incorporating changes (adding content, revising tone, tightening criteria). This pattern applies symmetrically to property schema definitions.

\section{Extended Experimental Results}
\label{app:extended-results}

\subsection*{Memory Density and Output Quality (E2)}

\begin{table}[H]
\centering
\small
\begin{tabular}{@{}lccc@{}}
\toprule
\textbf{Density} & \textbf{Avg Recalled} & \textbf{Score /100} & \textbf{Mem Use /30} \\
\midrule
Sparse (0 memories) & 0 & 69.3 & 19.0 \\
Minimal (3) & 3 & 86.0 & 28.0 \\
Light (7) & 7 & \textbf{88.0} & 28.5 \\
Moderate (12) & 12 & 84.4 & 26.0 \\
Rich (20) & 20 & 85.2 & 27.0 \\
Full (30) & 30 & \textbf{88.3} & 29.5 \\
\bottomrule
\end{tabular}
\end{table}

Zero entity memory produces measurably lower output quality (69.3/100). The first three memories provide a \textbf{+24\% relative quality jump} (69.3 $\to$ 86.0). Quality plateaus around \emph{light} density (7 memories, 88.0) with diminishing returns beyond, approximately 7 high-signal governed memories are sufficient to reach near-peak personalization quality in this evaluation setting.

\subsection*{Progressive Delivery Savings (E4)}

\begin{table}[H]
\centering
\small
\begin{tabular}{@{}llccc@{}}
\toprule
\textbf{Step} & \textbf{Task} & \textbf{Without} & \textbf{With} & \textbf{Savings} \\
\midrule
1 & Cold outreach & 9,312 & 9,312 & 0\% \\
2 & Follow-up w/ pricing & 6,802 & 959 & \textbf{85.9\%} \\
3 & Support escalation & 2,873 & 2,873 & 0\% \\
4 & Troubleshooting & 2,698 & 1,358 & \textbf{49.7\%} \\
5 & Closing proposal & 9,482 & 982 & \textbf{89.6\%} \\
\midrule
\textbf{Total} & & \textbf{31,167} & \textbf{15,484} & \textbf{50.3\%} \\
\bottomrule
\end{tabular}
\end{table}

Savings are topic-dependent: re-entrant steps (same governance domain already loaded) achieve 50--90\% savings; steps entering new domains require fresh context and save 0\%. The 50.3\% overall savings reflects a realistic mixed-domain workflow.

\subsection*{Write-Side Deduplication (E6)}

Across five overlapping sources for a single entity, the system stored \textbf{33 unique memories} while skipping \textbf{162 duplicates} (83.1\% dedup rate) with zero false positives. Near-miss facts (semantically similar but factually distinct) were correctly preserved.

\section{Schema Worked Example and Diagnostic Patterns}
\label{app:schema-example}

\subsection*{Worked Example: Per-Property Refinement}

\emph{Before refinement:}
\begin{lstlisting}
Property: "Technology Stack"
  type: text
  description: "The company's technology"
\end{lstlisting}

Phase~2 classifies this as \texttt{low\_confidence}, the description is too vague, producing inconsistent extraction.

\emph{After refinement:}
\begin{lstlisting}
Property: "Technology Stack"
  type: text
  description: "The primary technology infrastructure
    used by the company, including programming languages
    (e.g., Python, Java), frameworks (e.g., React, Django),
    cloud platforms (e.g., AWS, Azure), and databases
    (e.g., PostgreSQL, MongoDB). Focus on technical stack
    decisions rather than product or SaaS tool usage."
\end{lstlisting}

\subsection*{Diagnostic Patterns}

\begin{table}[H]
\centering
\caption{Recurring diagnostic patterns from evaluation records.}
\label{tab:diagnostic-patterns}
\small
\begin{tabular}{@{}p{3cm}p{4.5cm}p{4cm}@{}}
\toprule
\textbf{Score Pattern} & \textbf{Interpretation} & \textbf{Indicated Action} \\
\midrule
Low Context Util., high Completeness & Agent succeeded despite routing issues & Improve governance metadata \\
High Context Util., low Completeness & Appropriate context but insufficient memory & Improve memory coverage \\
Low Personalization, high Accuracy & Entity memories sparse or not recalled & Check density; review recall \\
Low across all criteria & Model or prompt issue & Review model \& system prompt \\
High variance within criterion & Schema-data alignment issue & Refine low-scoring types \\
\bottomrule
\end{tabular}
\end{table}

\subsection*{Evaluation Rubric Presets}

\begin{table}[H]
\centering
\small
\begin{tabular}{@{}ll@{}}
\toprule
\textbf{Preset} & \textbf{Criteria (weight)} \\
\midrule
Default & Accuracy (25), Relevance (25), Completeness (25), Context Util.\ (25) \\
Sales & Personalization (30), Value Prop.\ (25), CTA (20), Tone (25) \\
Support & Problem Understanding (25), Solution Accuracy (30), Clarity (25), Empathy (20) \\
Research & Thoroughness (30), Source Quality (25), Analysis (25), Organization (20) \\
\bottomrule
\end{tabular}
\end{table}

\end{document}